\documentclass[conference]{IEEEtran}
\IEEEoverridecommandlockouts
\usepackage{cite}
\usepackage{float}
\usepackage{amsmath,amssymb,amsfonts}
\usepackage{algorithmic}
\usepackage{graphicx}
\usepackage{textcomp}
\def\BibTeX{{\rm B\kern-.05em{\sc i\kern-.025em b}\kern-.08em
    T\kern-.1667em\lower.7ex\hbox{E}\kern-.125emX}}

\usepackage{multirow}
\usepackage{multicol}
\usepackage{booktabs} 
\usepackage[table]{xcolor}
\usepackage{subfigure}
\usepackage{todonotes}
\usepackage{xcolor}

\def\bs{\boldsymbol}

\title{Fraud Detection Using Optimized Machine Learning Tools Under Imbalance Classes}

\author{\IEEEauthorblockN{Mary Isangediok}
\IEEEauthorblockA{\textit{Department of Mathematics and Statistics} \\
\textit{Texas A\&M University-Corpus Christi}\\
Corpus Christi, USA \\
misangediok@islander.tamucc.edu}
\and
\IEEEauthorblockN{Kelum Gajamannage}
\IEEEauthorblockA{\textit{Department of Mathematics and Statistics} \\
\textit{Texas A\&M University-Corpus Christi}\\
Corpus Christi, USA \\
kelum.gajamannage@tamucc.edu}
}

\begin{document}

\maketitle
\thispagestyle{empty}
\pagestyle{empty}

\begin{abstract}
Fraud detection is considered to be a challenging task due to the changing nature of fraud patterns over time and the limited availability of fraud examples to learn such sophisticated patterns. Thus, fraud detection with the aid of smart versions of machine learning (ML) tools is essential to assure safety. Fraud detection is a primary ML classification task; however, the optimum performance of the corresponding ML tool relies on the usage of the best hyperparameter values. Moreover, classification under imbalanced classes is quite challenging as it causes poor performance in minority classes, which most ML classification techniques ignore. Thus, we investigate four state-of-the-art ML techniques, namely, logistic regression, decision trees, random forest, and extreme gradient boost, that are suitable for handling imbalance classes to maximize precision and simultaneously reduce false positives. First, these classifiers are trained on two original benchmark unbalanced fraud detection datasets, namely, phishing website URLs and fraudulent credit card transactions. Then, three synthetically balanced datasets are produced for each original data set by implementing the sampling frameworks, namely, random under sampler, synthetic minority oversampling technique (SMOTE), and SMOTE edited nearest neighbor (SMOTEENN). The optimum hyperparameters for all the 16 experiments are revealed using the method RandomzedSearchCV. The validity of the 16 approaches in the context of fraud detection is compared using two benchmark performance metrics, namely, area under the curve of receiver operating characteristics (AUC ROC) and area under the curve of precision and recall (AUC PR). For both phishing website URLs and credit card fraud transaction datasets, the results indicate that extreme gradient boost trained on the original data shows trustworthy performance in the imbalanced dataset and manages to outperform the other three methods in terms of both AUC ROC and AUC PR.
\end{abstract}


\section{Introduction}
Technological innovation has drastically altered how we interact with our surroundings through the internet. From simple tasks like browsing the web to complex tasks like diagnosing patients, the internet has highly influenced peoples' daily lives \cite{mcmillan2006coming, rice2006influences}. This surge in usage has unintentionally increased cybercrime and heightened concerns about potential fraud and cybersecurity dangers \cite{yeoh2019artificial}. Email spam is the method most frequently employed to help carry out these online frauds, in which botnets are utilized to send a large number of unwanted emails to naive recipients that contain malicious URLs or phishing websites \cite{barroso2007botnets}. These botnets are also used to distribute malicious software known as malware \cite{kigerl2021routine} to damage or exploit computers with the sole purpose of monitoring the activities of the victim. Some examples of malware are viruses, ransomware, Trojan horse, spyware, etc. \cite{tahir2018study}.

\begin{figure}[htp]
    \centering
    \includegraphics[width=\linewidth]{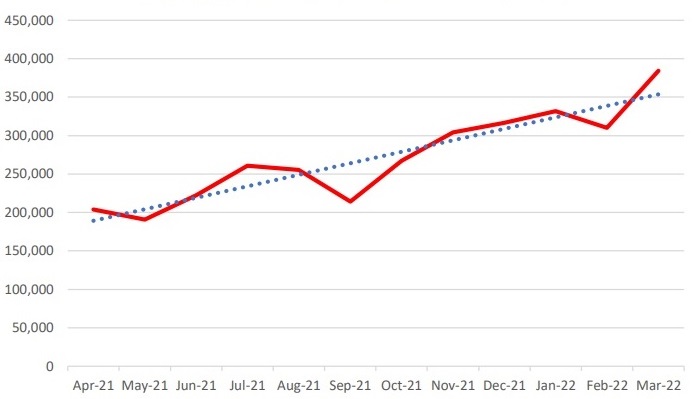}
    \caption{Phishing activities (red color) and its linear fit (dotted line), from the $2$-nd quarter of 2021 to the $1$-st quarter of 2022 available in Ref.~\cite{apwg}. The y-axis measures the total number of phishing attacks globally reported by APWG members and by members of the public.}
    \label{fig:phishing}
\end{figure}

Phishing is an illegal attempt to deceive people by employing social engineering and technological trickery to acquire their passwords, financial account information, and personal information \cite{khonji2013phishing, apwg}. The objective of phishing is to trick users into visiting a fake website by sending spam emails. The widespread use of social media and the increase in the sharing of personal information without any security measures has made phishing attempts increasingly common recently \cite{egele2013detecting}. Anti-Phishing Working Group, a not-for-profit organization dedicated to eradicating fraud and identity theft caused by phishing and related incidents, \cite{apwg} shows a steady rise in phishing activities with its highest observation in the first quarter of 2022 of over a million phishing attacks (see Fig.~\ref{fig:phishing}). The use of firewalls and antivirus software is a typical phishing protection technique since phishing assaults can result in identity theft and credit card fraud.

\begin{figure}[htp]
    \centering
    \includegraphics[width=\linewidth]{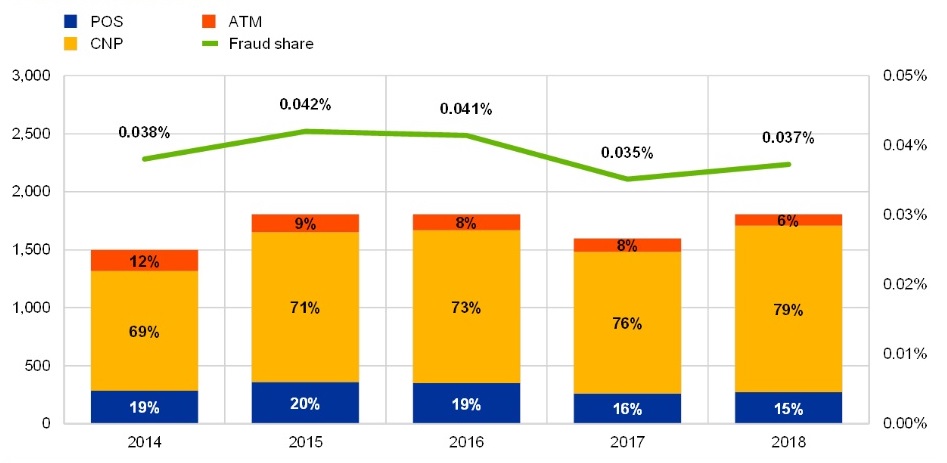}
    \caption{Total value of card frauds from 2014-2018 available in Ref.~\cite{ecb}. Left-hand scale: total fraud cost (in millions of Euros); right-hand scale: fraud cost as a \% of transaction cost (percentages). Card-not-present (CNP) accounted for a majority of the transaction report. The volume of automated teller machines (ATMs) and point-of-sale (POS) terminal frauds as a share of total fraud decreased in 2018 compared with 2017, whereas the volume of card-not-present (CNP) fraud as a share of total fraud increased.}
    \label{fig:card fraud}
\end{figure}

Every year, credit card fraud—a type of identity theft involving unauthorized use of another person's credit card information—leads to staggering financial losses \cite{nil20}. In $2018$, the total value of fraudulent transactions using cards issued within the Single Euro Payments Area, a European Union program for payment integration that simplifies euro-denominated bank transactions, amounted to €1.80 billion \cite{ecb}. By $2025$, gross fraud loss worldwide is projected to be about $\$35.31$ billion \cite{nil20}. A physical duplicate of a card is required for transactions made at ATMs and point-of-sale (POS) terminals under card-present (CP) schemes, whereas a card-not-present (CNP) scheme allows for transactions to be made over the phone, online, or by mail without the need for a physical copy of the card \cite{lereproducible}. CP payment methods have been in use for longer than CNP payment methods, and they are more resistant to fraudulent trends because they use cutting-edge technology like biometric identification and two-factor authentication to monitor and control CP fraud. To reduce the risk of CNP fraud, the authorization process for a transaction uses a dynamically changing visa card verification value. \cite{nil22}. Gross CP fraud losses from ATMs decreased by 1.8\% from 2018 \cite{nil20}, and according to the European Central Bank report of 2020, CNP fraud accounted for about 79\% of the total fraud transactions in 2018 \cite{ecb} as shown in Fig.~\ref{fig:card fraud}.

Robust systems to detect frauds would have great impacts on global finance through reduced fraud mitigation costs \cite{nil22}, increased customers' trust in merchants, and reduced victims' loss costs \cite{nil20}. Phishing and CNP fraud prevention are arguably challenging due to their volatility in fraud patterns over time as fraudsters take advantage of advances in technology (i.e., botnets, and malware) to exploit the vulnerabilities of computer users \cite{nil20}. Fraud detection systems are used to identify unusual behavior or pattern in users' transaction data. Historically, statistical methods such as regression analysis  \cite{mercer1990fraud}, discriminant analysis \cite{mahmoudi2015detecting}, and Markov chain Monte Carlo analysis \cite{abu2007comparison} have been applied to fraud detection. A major drawback however is that pure statistical tools are not suitable for adaptive learning and may not easily recognize the constantly evolving fraud patterns \cite{prakash2015optimized}.

Over the past few decades, a wide range of machine learning (ML) approaches has been employed for fraud detection \cite{sinayobye2018state, sadgali2018detection, priscilla2019credit, mittal2019performance}. Fraud detection is a type of anomaly detection where fraud observations are referred to as "anomalies" \cite{lereproducible}. Manifold learning frameworks as the ones presented in Refs.~\cite{gajamannage2019nonlinear, gajamannage2015identifying, gajamannage2015detecting}, are capable of accurately detecting anomalies of a diverse range of datasets, but these methods are not adequately tested on fraud detection. These anomalies can be detected using either supervised ML techniques such as autoencoders \cite{gajamannage2022reconstruction} or unsupervised ML approaches such as low-rank matrix completion \cite{gajamannage2021bounded}.
Logistic regression, decision trees, and naive Bayes are popular baseline models used in supervised learning owing to their simplicity and interpretability \cite{lereproducible}. Rule-based expert systems, k-nearest neighbor, and Support Vector Machine (SVM) produce low precision scores \cite{mittal2019performance} and require more training time \cite{priscilla2019credit}. The hidden Markov model (HMM) is used in many fraud detection systems due to its ability to find unknown or hidden patterns in a skewed dataset.  Ref.~\cite{prakash2015optimized} proposed an optimized semi-hidden Markov model to improve model performance nevertheless, HMM and its optimized variants are computationally complex \cite{priscilla2019credit}. Ensemble learning has proven to be very effective and versatile in automated decision-making systems \cite{lereproducible}. Random forests and boosting methods are able to avoid overfitting the training data \cite{priscilla2019credit}; however, they can easily become complex with many trees and require a lot of computation time. Fraud detection systems that rely on the flexibility of artificial neural networks require a large amount of training data to learn patterns and successfully detect outliers \cite{chang2007intrusion}. These models are however computationally expensive and are susceptible to overfitting \cite{priscilla2019credit}. Hybrid techniques which combine statistical methods like discriminant analysis, bayesian, and HMM to neural networks have shown great promise in minimizing the misclassification of fraudulent observations  \cite{maes2002credit}.  Isolation forest and local outlier factor are unsupervised methods that can easily find new and unusual patterns in a dataset and have shown great performance in fraud detection, but they can become computationally complex and require a large amount of data to learn from \cite{mittal2019performance}.

A common challenge that becomes apparent with ML fraud detection systems is that the limited amount of fraud observations causes a large discrepancy between the number of observations in each class, making it difficult for classifiers to learn predictive information from the highly skewed dataset \cite{fernandez2018learning, chawla2009data}, which can lead to a bias towards the majority class (genuine observations). A standard practice to address class imbalance in datasets is the use of four common strategies \cite{5978225}: algorithm-level approach, data-level approach, cost-sensitivity learning approach, and ensemble approach \cite{lereproducible, ling2008cost}. In order to account for the importance of the minority class, the algorithm-level approach develops new algorithms or alters those that already exist \cite{wu2005kba}. Cost-sensitivity learning considers the misclassification costs by modifying the optimization function in the training step of the learning algorithm \cite{lereproducible}. For instance, a classifier may give false negatives a higher cost than false positives, emphasizing any right classification or mistake relating to the positive class \cite{ali2013classification}. In the data-level approach, the class distribution of the dataset is rebalanced by adding a preprocessing step before the training algorithm is implemented \cite{batista2004study}. To do this, the imbalance ratio in training data is decreased by either using under-sampling or over-sampling methods\cite{ali2013classification}. Under-sampling eliminates fewer data instances from the majority class, whereas over-sampling duplicates data instances from the minority class \cite{JMLR}. Ensemble approaches are intended to boost the accuracy of a single classifier by training many classifiers and merging their decisions to produce a single class label \cite{rokach2010ensemble}. These strategies aim to skew the decision boundaries in favor of the minority class \cite{lereproducible}.

There is great potential to advance fraud detection using ML techniques; thus, it is worthwhile to carefully investigate the capabilities of hybrid frameworks for fraud detection. This paper studies the effect of 3 resampling approaches: RandomUnderSampler (RUS) \cite{lereproducible}, Synthetic Minority Oversampling Technique (SMOTE) \cite{chawla2002smote}, and a combination of oversampling--SMOTE and undersampling--Edited Nearest Neighbor (ENN), denoted as SMOTEENN. These data-level resampling strategies are trained on 4 optimized ML classifiers, namely, logistic regression, decision tree, random forest, and extreme gradient boost. 

To describe the performance of these four classifiers more accurately, two performance metrics were investigated, namely area under the curve of receiver operating characteristics (AUC ROC) and area under the curve of precision and recall (AUC PR). This experiment attempts to increase fraud detection and reduce the number of legitimate observations that are incorrectly labeled as frauds. Resampling strategies' potential to create classifiers with a superior AUC ROC was demonstrated through a comparison of all methodologies. The investigation also revealed that extreme gradient boost is robust to class imbalance. 

The rest of this paper is organized as follows. Section II outlines the investigated ML approaches. Section III describes the data and experimental setup. Section IV reports the details of the experimental results and discussion about the comparative analysis. We conclude in Section V and suggest future areas of research. Table~\ref{tab:abr} presents all the abbreviations used in this paper. 
\begin{table}[htp]
\centering
\caption{Abbreviations used in this paper and their descriptions.}
\begin{tabular}{p{1.4cm}|p{6.5cm}}
\toprule
Abbreviation & Description\\
 \midrule
  AUC ROC & Area Under the Curve of Receiver Operating Characteristics\\
  AUC PR & Area Under the Curve of Precision and Recall \\
  FN & False Negative \\
  FP & False Positive \\
  FPR & False Positive Rate \\
  TN & True Negative \\
  TP & True Positive \\
  TPR & True Positive Rate \\
  LR & Logistic Regression \\
  DT & Decision Tree \\
  RF & Random Forest \\
  XGB & Extreme Gradient Boosting (XGBoosting) \\
  orig & original imbalanced dataset \\
  RUS & RandomUnderSampler \\
  SMOTE & Synthetic Minority Oversampling Technique \\
  ENN & Edited Nearest Neighbor \\
  SMOTEENN & SMOTE + ENN \\
  
\bottomrule
\end{tabular}
\label{tab:abr}
\end{table}
\section{Methods}
Methods include, 1) four classifiers that are used to classify fraud and genuine data instances; and 2) two metrics that we use to assess the performance of the classifiers. 

\subsection{Classifiers}
Here, we present four classifiers logistic regression, decision tree, random forest, and extreme gradient boosting (or XGBoost), that are used in our analysis.

\subsubsection{Logistic Regression}
Logistic regression (LR) is the most widely used learning algorithm for binary classification tasks because of its simplicity \cite{sperandei2014understanding}. Taking the fraud observations as $1$'s and the genuine observations as $0$'s, LR models the probability of a transaction being identified as fraudulent given other features in the dataset at a specified threshold \cite{sperandei2014understanding}. If the probability is greater than the threshold it is fraud else, non-fraud \cite{sperandei2014understanding}. Let, $\bs{x}\in\mathbb{R}^m$ denote the input feature vector of length $m$, then the response $z$ is given as a straight line $z = \bs{w} \cdot \bs{x} +b$, where $\bs{w}$ is the weights and $b$ is the bias term estimated during training. Thus, the logistic function is given as 
\begin{equation}
g(z) = \frac{1}{1+e^{-z}}, \quad 0<g(z)<1.
\end{equation}
Then, the LR model given the probability $f_{\bs{w},b}(\bs{x})$ is
\begin{equation}
f_{\bs{w},b}(\bs{x}) = \frac{1}{1+e^{-(\bs{w} \cdot\bs{x} +b)}}, \quad 0<p(z)<1.
\end{equation}
For threshold $t$, probability corresponding to the positive class $1$ is interpreted as 
$f_{\bs{w},b}(\bs{x}) = P(y=1|\bs{x};\bs{w},b)$. LR is reliable and requires less computation time for the training phase. However, for datasets with extreme class imbalance, this supervised learning approach is less likely to produce competitive outcomes \cite{priscilla2019credit}.

\subsubsection{Decision Tree}
Decision tree (DT) has a hierarchical tree structure that consists of a root node, branches, internal nodes, and leaf nodes \cite{wu2008top}. DT chooses what feature, i.e., $x_{i}$'s, to split at a node based on what choice of feature reduces entropy. Let, $p_{1}$ be the fraction of the data instances that belong to the first class of interest (i.e., fraudulent) in a binary classification, then the entropy $H(p_{1})$ is defined as
\begin{equation}
    H(p_{1}) = - p_{1}log_{2}(p_{1}) - (1-p_{1})log_{2}(1-p_{1}).
\end{equation}
A stump of a decision tree is defined as a parent node and its two children and information gain is a function, computed at each stump, that is defined as the entropy of a parent minus the entropy of its candidate splits, i.e., children. This recursive algorithm is trained with the data so that the DT model estimates the best splitting while maximizing the total information gain. We train a DT with unlimited depth and min\_samples\_split=2 so that training continues until all leaves are pure or until all leaves contain less than min\_samples\_split instances. Like LR, DTs are simple to build and easy to understand however, they can easily become complex with more depths and require more computation time for the training phase \cite{priscilla2019credit}.

\subsubsection{Random Forest}
Given a training set, random forest (RF) is an ensemble of decision trees, where each tree in the ensemble is comprised of a data sample drawn from the training set repeatedly, say $k$ times, with replacement \cite{breiman2001random}. This creates a random forest for the dataset and then makes classifications using RFs. Out-of-bag sampling is used to give ongoing estimates of the generalization error of the combined ensemble of trees. The model prediction is made by taking the majority vote from all classification trees. RFs give better predictions when compared with a single model and are less likely than DTs to overfit during training but this classifier's processing time increases with complexity.

\subsubsection{Extreme gradient boosting (or XGBoosting)}
Extreme gradient boosting (XGB) \cite{chen2016xgboost}, improves predictive power by building an ensemble of trees that uses sampling with replacement to create new training sets. Instead of picking each data instance with equal probability as with RFs, XGB makes it more likely to pick instances that the previously trained tree misclassified. Let, $\bs{x}_i\in\mathbb{R}^m$ is a vector of $m$ features, the tree ensemble model of interest uses $K$ additive functions to predict the output $\hat{y}_i$ as
\begin{equation}
    \hat{y}_{i} = \sum_{k=1}^{K} f_{k}(\bs{x}_i), \quad f_{k}\in\mathcal{F}
\end{equation}
Where $\mathcal{F}=\{f_k\vert k = 1, \dots, K\}$ is the space of regression trees and each $f_{k}$ corresponds to an independent tree structure, say $q$and leaf weights $w$ \cite{chen2016xgboost}. If $\hat{y}_i^{(t)}$ is the prediction of the $i$-th instance of the $t$-th iteration, we greedily add $f_t$ to minimize the objective $L$ such that,
\begin{equation}
    L^{(t)} = \sum_{i=1}^{n}l(y_{i},\hat{y}_{i}^{(t-1)} +f_t(\bs{x}_i)) + \Omega(f_{t}),
\end{equation}
where $\Omega(f)=\gamma T+1/2\lambda \|w\|^2$. Here, $l(f,g)$ is a differentiable convex loss function that measures the difference between $f$ and $g$, $T$ is the number of leaves in the
tree, and $\gamma$ and $\lambda$ are constants. During the training process, this objective $L^{(t)} $ is optimized using a second-order approximation of it for a fixed tree structure $q$ and estimates the weights $w$ of the leaves. \cite{chen2016xgboost}

The algorithm starts with one weak learner and iteratively adds new weak learners to approximate functional gradients. The final ensemble model is constructed by a weighted summation of all weak learners. The parameter scale\_pos\_weight controls the balance of positive and negative weights. We keep this parameter relatively small as  $\frac{\sqrt{\sum_{i=1}^{n}y}}{\sum_{i=1}^{n}\hat{y}}$ for credit card dataset since its classes are extremely imbalanced. XGB contains a regularization term which helps it to avoid overfitting. The hyperparameters of XGB can be tuned to account for class imbalance, making it an effective learning algorithm for skewed datasets. However, high computation time is the biggest disadvantage of this algorithm.

\subsection{Performance metrics}
We assess the performance of each of the four classifiers based on two performance metrics, AUC ROC and AUC PR. There are four possible outcomes of a classifier: 1) classifying a fraudulent instance as fraudulent, this is named a true positive (TP); 2) classifying a fraudulent instance as genuine, this is named a false positive (FP); 3) classifying a genuine instance as genuine, this is named as true negative (TN); and 4) classifying a genuine instance as fraudulent, this is named as false negative (FN) \cite{lereproducible}. The confusion matrix of binary classifiers is given in Table~\ref{tab:confmat}.
\begin{table}[htp]
\centering
\caption{Confusion Matrix}
\begin{tabular}{l l l l}
\cmidrule{3-4}
\multicolumn{1}{c}{} & & \multicolumn{2}{c}{Predicted Class} \\
\cmidrule{3-4}
 & & \textbf{\emph{Genuine (0)}} & \textbf{\emph{Fraud (1)}} \\
\midrule
\multirow{2}{*}{Actual Class} & \textbf{\emph{Genuine (0)}} & True Negative(TN) & False Positive(FP) \\
\cmidrule{2-4}
 & \textbf{\emph{Fraud (1)}} & False Negative(FN) & True Positive(TP) \\
\bottomrule
\end{tabular}
\label{tab:confmat}
\end{table}

\subsubsection{Area Under the Curve of Receiver Operating Characteristics}
The receiver operating characteristic (ROC) curve \cite{lereproducible}, is especially useful with skewed class distribution because they are insensitive to changes in class distribution. The ROC curve is obtained by plotting true positive rate (TPR), also called \emph{recall}, i.e.,  $\frac{TP}{TP + FN}$, against the false positive rate (FPR), i.e., $\frac{FP}{FP + TN}$, for all possible fraud probability values returned by a classifier at different thresholds \cite{lereproducible}. Area under the curve of receiver operating (AUC ROC) of a classifier is a scalar value equivalent to the probability that the classifier will rank a randomly chosen positive instance higher than a randomly chosen negative instance \cite{lereproducible}.

\subsubsection{Area Under the Curve of Precision and Recall}
The precision-recall (PR) curve \cite{lereproducible} is created by graphing the precision, i.e., $\frac{TP}{TP + FP}$, against the recall for each conceivable fraud probability value returned by a classifier at the different thresholds. The primary benefit of the PR curve is the inclusion of evidence classifiers with high recall and high precision \cite{lereproducible}. Ref.~\cite{saito2015precision} states that high TPR and low FPR values are significant in a fraud detection problem therefore, the precision-recall curve is an essential performance measure for imbalanced datasets. Area under the curve of precision and recall (AUC PR) of a given classifier is a single number summary of the information in the precision-recall (PR) curve, calculated by computing the average precision at each threshold, with the increase in recall from the previous threshold used as the weight. The average precision of a random classifier decreases as the class imbalance ratio increases, this property makes AUC PR better reflect the challenge related to a class imbalance problem.

\section{Data and Experimental Setup}

\subsection{Data description}
Our fraud detection study is performed on two datasets, namely, phishing website URLs and credit card transactions.

\begin{table}[htp]
\caption{Datasets with proportion of negative and positive classes}
\resizebox{\linewidth}{!}
{\begin{tabular*}{\columnwidth}{ @{\extracolsep{\fill}}l l l l}
    \toprule
    Dataset & Total & Genuine & Fraud \\
    \midrule
    Phishing website URLs & 88647 & 50000 & 30647 \\
    Credit card fraud & 284807 & 284315 & 492 \\
    \bottomrule
\end{tabular*}}
\label{Data}
\end{table}

\begin{itemize}
    \item \textbf{Phishing website dataset.} This dataset contains $111$ features that are extracted from the collections of website URLs with $88,647$ entries and no missing values \cite{VRBANCIC2020106438}. \emph{Phishing} is the binary response variable to predict, where $0$ represents genuine sites and $1$ represents phishing/fraudulent sites. $34.57\%$ of the total data are phish website instances.
\end{itemize}

\begin{itemize}
    \item \textbf{Credit card fraud dataset.} The highly unbalanced dataset provided by the Université Libre de Bruxelles (ULB) ML group \cite{kag1} contains $0.17\%$ fraudulent transactions out of the $284,807$ transactions. This is a 30-feature dataset that contains credit card transactions that occurred in two days made by European cardholders. 
\end{itemize}

\subsection{Preprocessing}
The quality of data is vital in every ML algorithm; therefore, it is essential to remove redundant features and increase efficiency. The features \emph{Time} and \emph{Amount} of the credit card dataset are normalized by z-scoring. Decision trees and tree ensembles are robust to scale and distribution of values in a dataset and generally do not require normalization; however, the normalization helps to minimize the computation time of each model. Features of the phishing site dataset are directly used without normalization because the data distribution is relatively small.

\subsection{Proposed resampling techniques}
In this paper, we compile three synthetically balanced datasets from the two imbalanced datasets. To implement this, we have utilized an imbalanced-learn \cite{JMLR} library that includes resampling techniques such as RUS, SMOTE, and SMOTEENN. RUS aims to balance class distribution by randomly eliminating samples from the majority class \cite{lereproducible}. Some data instances of the majority class in the training set are randomly undersampled to balance the dataset. This approach is simple and computationally efficient; but it can cause loss of valuable information needed for better predictive performance of the model. The number of genuine observations in our datasets are resampled so that the total number of genuine cases is equal to the total number of fraud/phishing cases. SMOTE is an oversampling technique that generates synthetic data instances of the minority class by interpolation \cite{chawla2002smote}. It creates a probability distribution to describe the smaller class and enlarges the decision boundary to encompass nearby minority class examples, reducing the risk of misclassifying the minority class. SMOTE can become computationally expensive with larger datasets. SMOTEENN is a resampling technique that combines over and undersampling approaches \cite{batista2003balancing}. Here, SMOTE is first applied to the training set to oversample the minority class. All data examples of each class that diverge from their neighborhood is eliminated by edited nearest neighbour (ENN); therefore, this technique is able to clean up more noise after oversampling may not produce equal instances for each class as in RUS and SMOTE.

\subsection{Evaluation Pipeline}
Stratified sampling by class helps to maintain the same level of imbalance in the train and test sets. The dataset is distributed in the ratio of 80\% for training and 20\% for  testing. Our evaluation pipeline consists of three parts: 
\begin{enumerate}
\item RandomizedSearchCV is used to obtain optimal parameters for each classifier with stratified $k$-fold cross-validation. The two different datasets are partitioned by stratified split into 10 equal-size validation pairs and the class proportion across each pair is maintained. To reduce the effect of randomness for the 10-fold split, we set the number of iterations to 5. Table II contains the list of the tuned parameter values used to optimize the performance of each classifier for both datasets.
\item Now that we have obtained the optimal sets of hyperparameters for training, we create three new synthetically balanced datasets with RUS, SMOTE, and SMOTEENN.
\item We then train the four classifier models using the original unbalanced dataset and the three synthetically balanced datasets respectively.
\end{enumerate}

\begin{table*}[htp]
\small\centering
\caption{Optimal hyperparameter sets for the datasets phishing website URLs and credit card transactions.}
\begin{tabular}{l l l l l}
    \toprule
    Dataset & LR & DT & RF & XGB \\
    \midrule
    Phishing websites URLs& solver = liblinear & criterion = entropy & n\_estimators = 100 & n\_estimators = 200 \\
     & penalty = $l_2$ & & max\_depth = 20 & max\_depth = 15 \\
     & max\_iter = 500 & & oob\_score = True & colsample\_bytree = 0.9 \\
     & C = 5 & & warm\_start = True & scale\_pos\_weight = ratio \\
     & & & & learning\_rate = 0.1 \\
     & & & & gamma = 0.3 \\
     \hline
    Credit card fraud& solver = lbfgs & criterion = entropy & n\_estimators = 50 & n\_estimators = 100 \\
     & penalty = $l_2$ & & max\_depth = 12 & max\_depth = 11 \\
     & max\_iter = 100 & & oob\_score = True & colsample\_bytree = 0.9 \\
     & C = 3 & & warm\_start = True & scale\_pos\_weight = ratio \\
     & & & & learning\_rate = 0.1 \\
     & & & & gamma = 0.3 \\
    \bottomrule
\end{tabular}
\label{Hyperparameter Tuning}
\end{table*}

\begin{table*}[htp]
\caption{Performance comparison of 16 experiments: four original classifiers, namely, logistic regression (LR), decision tree (DT), random forest (RF), and, XGB (XGB); and each classifier with three sampling techniques, namely, RandomUnderSampler (RUS), Synthetic Minority Oversampling Technique (SMOTE), and SMOTE-Edited Nearest Neighbor (SMOTEENN). The best value for each metric among all the 16 experiments is colored in red }
\centering
\begin{tabular}{l|p{4mm} p{4mm} p{4mm} p{7mm} p{7mm} p{7mm}|p{4mm} p{4mm} p{4mm} p{7mm} p{7mm} p{7mm}}
\toprule
\multirow{3}{*}{Model} & \multicolumn{6}{c|}{Phishing website URLs dataset} &  \multicolumn{6}{c}{Credit card fraud dataset}\\
\cmidrule{2-13}
 & FP & FN & Recall & Precision & AUC ROC & AUC PR & FP & FN & Recall & Precision & AUC-ROC & AUC-PR \\
\midrule
 LR+orig & $760$ & $570$ & $0.91$ & $0.88$ & $0.976$ & $0.950$ & $43$ & $8$ & $0.87$ & $0.56$ & $0.989$ & $0.762$ \\
 LR+RUS & $1100$ & $360$ & $0.94$ & $0.84$ & $0.976$ & $0.950$ & $1800$ & $6$ & $0.94$ & $0.05$ & $0.990$ & $0.665$ \\
 LR+SMOTE & $1000$ & $360$ & $0.94$ & $0.85$ & $0.976$ & $0.950$ & $7$ & $1300$ & $0.93$ & $0.07$ & $0.992$ & $0.707$  \\
 LR+SMOTEENN & $1200$ & $420$ & $0.93$ & $0.83$ & $0.976$ & $0.950$ & $1300$ & $8$ & $0.92$ & $0.06$ & $0.985$ & $0.712$ \\
 
 DT+orig &$370$ & $420$ & $0.93$ & $0.94$ & $0.950$ & $0.900$ & $19$ & $22$ & $0.78$ & $0.80$ &$0.877$ &$0.589$ \\
 DT+RUS & $560$ & $310$ & $0.95$ & $0.91$ & $0.950$ & $0.900$ & $5400$ & ${\color{red}5}$ & ${\color{red}0.95}$ & $0.02$ & $0.919$ & $0.015$ \\
 DT+SMOTE & $400$ & $380$ & $0.94$ & $0.94$ & $0.950$ & $0.900$ & $99$ & $21$ & $0.79$ & $0.44$ & $0.892$ & $0.303$ \\
 DT+SMOTEENN & $480$ & $370$ & $0.94$ & $0.92$ & $0.950$ & $0.900$ & $152$ & $19$ & $0.81$ & $0.35$ & $0.902$ & $0.309$ \\
 
 RF+orig & $270$ & $240$ & $0.96$ & ${\color{red}0.96}$ & $0.995$ & $0.992$ & $4$ & $22$ & $0.78$ & $0.95$ & $0.983$ & $0.818$ \\
 RF+RUS & $480$ & $160$ & $0.97$ & $0.92$ & $0.995$ & $0.992$ & $200$ & $7$ & $0.93$ & $0.04$ & $0.979$ & $0.724$ \\
 RF+SMOTE & $340$ & $200$ & $0.97$ & $0.95$ & $0.995$ & $0.992$ & $35$ & $16$ & $0.84$ & $0.70$ & ${\color{red}0.995}$ & $0.814$ \\
 RF+SMOTEENN & $410$ & $200$ & $0.97$ & $0.94$ & $0.995$ & $0.992$ & $33$ & $14$ & $0.86$ & $0.72$ & $0.992$ & $0.684$ \\
 
 XGB+orig & ${\color{red}230}$ & $200$ & $0.97$ & ${\color{red}0.96}$ & ${\color{red}0.996}$ & ${\color{red}0.994}$ & ${\color{red}2}$ & $19$ & $0.81$ & ${\color{red}0.98}$ & $0.992$ & $0.849$ \\
 XGB+RUS & $360$ & ${\color{red}140}$ & ${\color{red}0.98}$ & $0.94$ & ${\color{red}0.996}$ & ${\color{red}0.994}$ & $2100$ & $6$ & $0.94$ & $0.04$ & $0.992$ & $0.737$ \\
 XGB+SMOTE & $270$ & $180$ & $0.97$ & ${\color{red}0.96}$ & ${\color{red}0.996}$ & ${\color{red}0.994}$ & $24$ & $15$ & $0.85$ & $0.78$ & $0.990$ & ${\color{red}0.851}$ \\
 XGB+SMOTEENN & $370$ & $210$ & $0.97$ & $0.94$ & ${\color{red}0.996}$ & ${\color{red}0.994}$ & $29$ & $15$ & $0.85$ & $0.74$ & $0.991$ & $0.817$  \\
\bottomrule
\end{tabular}
\label{Results}
\end{table*}

\begin{figure*}[htp]
\centering
 \subfigure{\includegraphics[width=6cm]{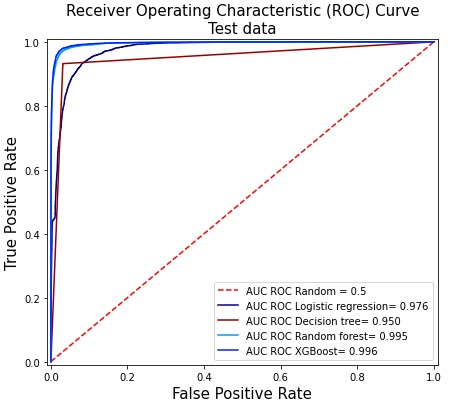}}
 \subfigure{\includegraphics[width=6cm]{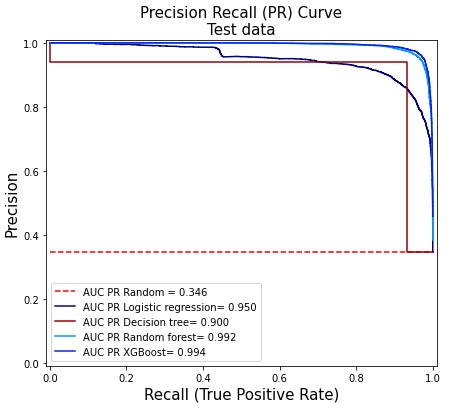}}
 \caption{AUC ROC and AUC PR of the four classifiers applied to the original phishing website URLs dataset.} 
 \label{fig:orig phishing}
\end{figure*}

\begin{figure*}[htp]
\centering
 \subfigure{\includegraphics[width=6cm]{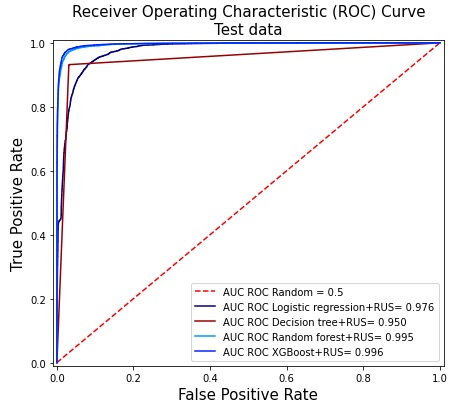}}
 \subfigure{\includegraphics[width=6cm]{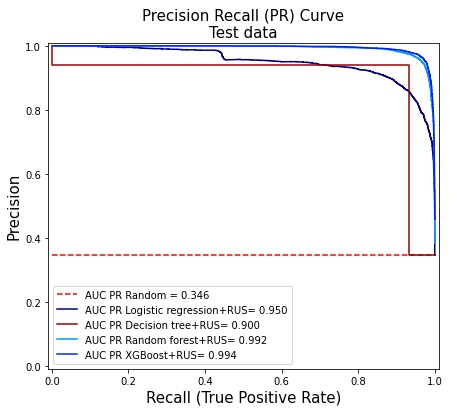}}
 \caption{AUC ROC and AUC PR of the four classifiers applied to the undersampled (RUS) phishing website URLs dataset.} 
 \label{fig:rus phishing}
\end{figure*}

\begin{figure*}[htp]
\centering
 \subfigure{\includegraphics[width=6cm]{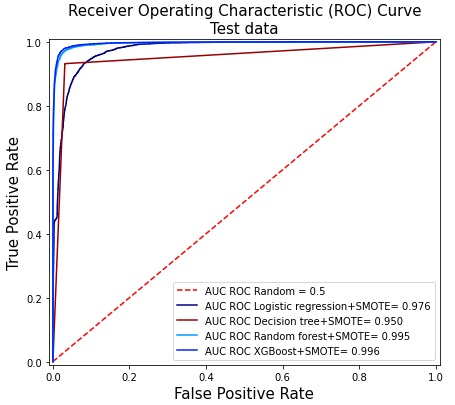}}
 \subfigure{\includegraphics[width=6cm]{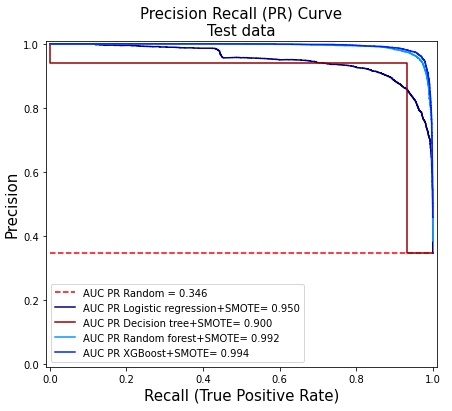}}
 \caption{AUC ROC and AUC PR of the four classifiers applied to the oversampled (SMOTE) phishing website URLs dataset.} 
 \label{fig:smote phishing}
\end{figure*}

\begin{figure*}[htp]
\centering
 \subfigure{\includegraphics[width=6cm]{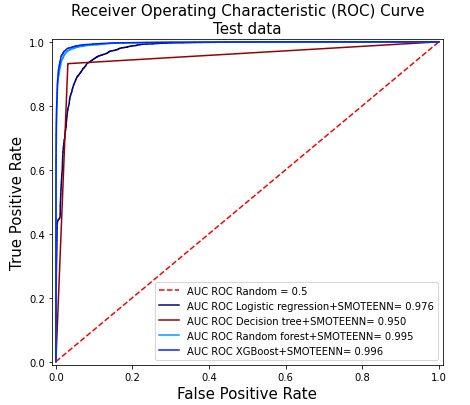}}
 \subfigure{\includegraphics[width=6cm]{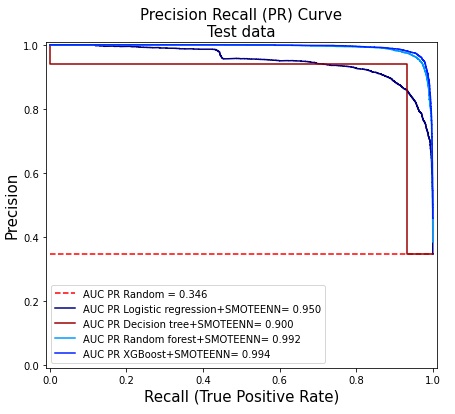}}
 \caption{AUC ROC and AUC PR of the four classifiers applied to the resampled (SMOTEENN) phishing website URLs dataset.} 
 \label{fig:smoteenn phishing}
\end{figure*}

\begin{figure*}[htp]
\centering
 \subfigure{\includegraphics[width=6cm]{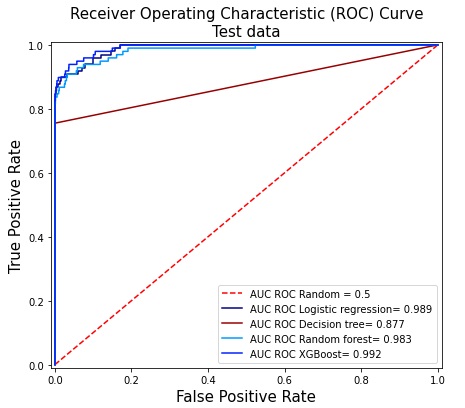}}
 \subfigure{\includegraphics[width=6cm]{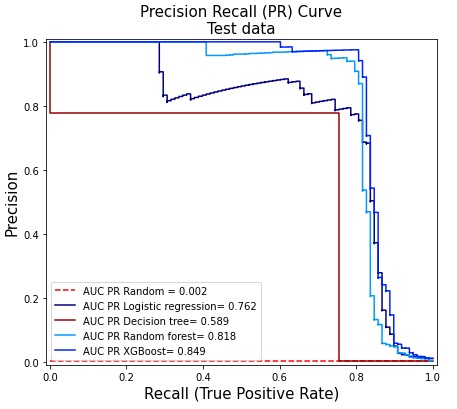}}
 \caption{AUC ROC and AUC PR of the four classifiers applied to the original credit card fraud dataset.} 
 \label{fig:orig creditcard}
\end{figure*} 

\begin{figure*}[htp]
\centering
 \subfigure{\includegraphics[width=6cm]{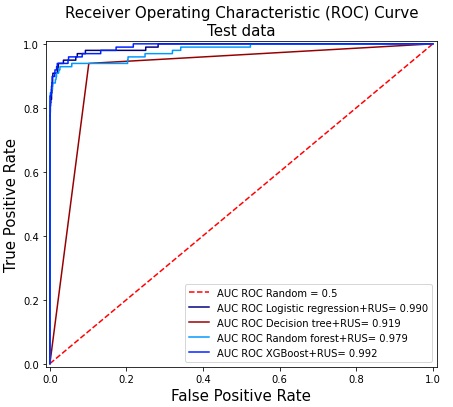}}
 \subfigure{\includegraphics[width=6cm]{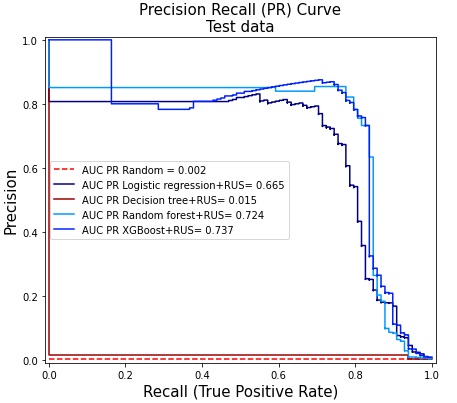}}
 \caption{AUC ROC and AUC PR of the four classifiers applied to the undersampled (RUS) credit card fraud dataset.} 
 \label{fig:rus creditcard}
\end{figure*}

\begin{figure*}[htp]
\centering
 \subfigure{\includegraphics[width=6cm]{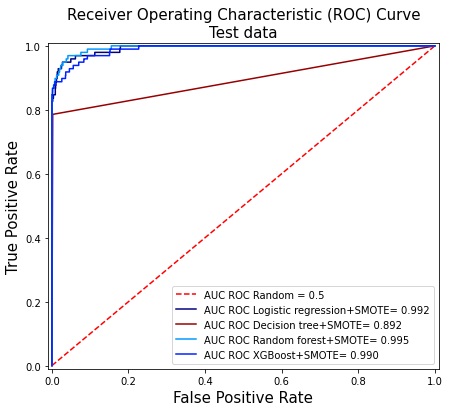}}
 \subfigure{\includegraphics[width=6cm]{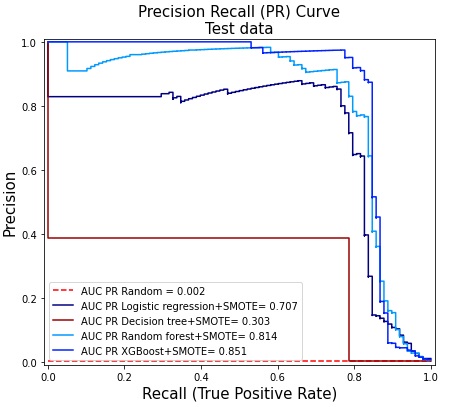}}
 \caption{AUC ROC and AUC PR of the four classifiers applied to the oversampled (SMOTE) credit card fraud dataset.} 
 \label{fig:smote creditcard}
\end{figure*}

\begin{figure*}[htp]
\centering
 \subfigure{\includegraphics[width=6cm]{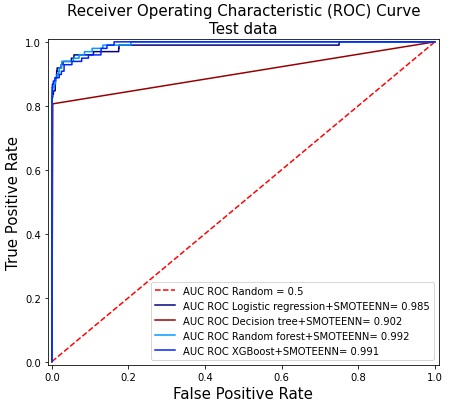}}
 \subfigure{\includegraphics[width=6cm]{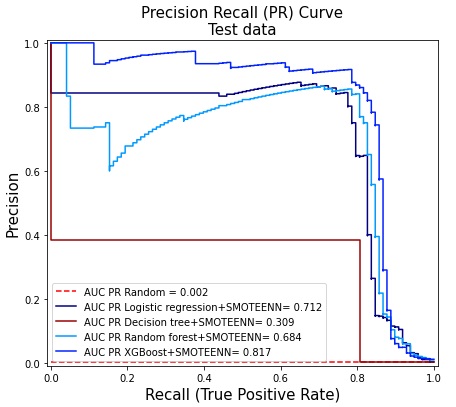}}
 \caption{AUC ROC and AUC PR of the four classifiers applied to the resampled (SMOTEENN) credit card fraud dataset.} 
 \label{fig:smoteenn creditcard}
\end{figure*}

\section{Results}
An efficient fraud detection system aims at optimizing recall and precision. Our study is based on 16 classifiers: four state-of-the-art classifiers where each of them is implemented one time as it is and three more times with resampling techniques, RUS, SMOTE, SMOTEENN. We compare the fraud detection performance of these 16 classifiers applied to two datasets, phishing website URLs and credit card fraud, with respect to two performance metrics, AUC ROC and AUC PR, and discuss the impact of these models. Table~\ref{Results} lists the performance metrics, FP, FN, recall, precision, AUC ROC, and AUC PRC of all 16 models. Figs.~\ref{fig:orig phishing}--\ref{fig:smoteenn creditcard} show the performance, with respect to AUC ROC and AUC PR, of all models on the two original imbalanced datasets. Figs.~\ref{fig:orig phishing}--\ref{fig:smoteenn phishing} provide evidence that four XGB models over perform other techniques when applied to the phishing dataset as XGB models gain $0.996$ and $0.994$ for AUC ROC and AUC PR, respectively. By identifying more instances of fraud and lowering false negatives, resampling techniques marginally enhance the performance of each classifier although increased false positives are observed. For the phishing dataset, the experiment that we train XGB on the original imbalance dataset outperformed all other approaches.

For the credit card dataset, random forest along with SMOTE performed better, in terms of AUC ROC, than other approaches. However, in terms of AUC PR, random forest with SMOTE performs marginally better than XGB with original imbalance credit card dataset. XGB with the original credit card dataset had both the best precision score of $0.98$ and the fewest false positives of 2. It is evidenced that the decision tree along with RUS can identify more instances of frauds as we observed the best recall of $0.95$. Models trained with resampled data appear to favor the minority class by maximizing recall. Classifiers trained on RUS as shown in Figs.~\ref{fig:rus phishing} and \ref{fig:rus creditcard} appear to provide the fewest FNs but also yield the most FPs, suggesting that RUS may not be a good technique for real-world fraud detection systems since it produces too many FPs. It is possible that SMOTEENN is ineffective for real-world fraud detection systems since it performs poorly in these datasets and consumes a lot of processing time. In comparison to RUS and SMOTEENN, classifiers trained on SMOTE exhibit a significantly higher recall and AUC ROC. With AUC ROC, AUC PR of phishing website URLs, and credit card data interpreted as $0.996$, $0.994$, and $0.992$, $0.849$ respectively, XGB is the most successful classification algorithm for minimizing false positives and maximizing recall among all methods.

\section{Conclusion and Future Work}
Fraud detection with the aid of smart techniques is essential to assure safety as modern humans are prone to cyber attacks with the surge of internet-based lifestyles. This study is based on detecting frauds by combining several families of machine learning techniques, namely, classification tools, resampling tools for imbalance classes, hyperparameter tuning frameworks. Especially, this research paper studied the use of four classifiers, namely, logistics regression (LR), decision tree (DT), random forest (RF), and extreme gradient boosting (XGB), along with three imbalance learning techniques, namely, random under sampler (RUS), synthetic minority oversampling technique (SMOTE), and SMOTE edited nearest neighbor (SMOTEENN), in fraud detection systems. These techniques  are implemented on two datasets, phishing website URLs and credit card frauds, after tuning the hyperparameters by RandomizedSearchCV.

The genuine class outweighs any fraudulent class in the majority of cybersecurity datasets. Using the imbalance learn library can help generate more distinct decision boundaries for each class since these classes differ significantly in several attributes. We found that the resampling techniques minimized FPR and AUC PR while optimizing AUC ROC. An ideal fraud detection system should utilize the models that minimize FPR and maximize precision because an operational fraud detection system typically works with much larger datasets on a regular basis and with a small team of investigators who manually check all false positives (genuine instances that are incorrectly flagged) \cite{lereproducible}.

In general, the algorithms XGB and random forest outperformed logistic regression and decision tree. In terms of precision and AUC PR, XGB outperformed random forest and showed more robustness to data imbalance. The AUC ROC of the classifiers trained on resampled datasets showed a small improvement with a moderate computational time; however, the classifiers work along with RUS attain a significant reduction in computation time. Additionally, metrics like precision and AUC PR are often negatively impacted by resampling techniques. Although an extensive hyperparameter tuning with GridSerchCV \cite{lereproducible}, could help with model selection, the computation time necessary to acquire the ideal values poses significant barrier to effectively exploring models' hyperparameters.

Future work will explore the potential of using neural networks based techniques, along with resampling and hyperparameter tuning, for fraud detection. It would include building a hybrid convolutional neural network and long short-term memory (CNN-LSTM) framework. CNNs are capable of learning feature representations without a need for feature engineering \cite{lereproducible}, and LSTMs are known for their feedback connections capable of learning long-term dependencies \cite{gajamannage2022real}. Combining these two powerful deep learning algorithms could result in significant performance when compared with RF and XGB. We will also investigate the performance of this framework when applied with imbalanced learning techniques for fraud detection.


\bibliographystyle{plain}

\end{document}